%% file: paper.tex
\documentclass[journal,twoside,print]{ieeecolor}
\usepackage{tmi}
\usepackage{cite}
\usepackage{amsmath,amssymb,amsfonts}
\usepackage{graphicx}
\usepackage{textcomp}
\usepackage{tikz}
\usetikzlibrary{arrows,calc,fit}
\usepackage{amsmath}
\usepackage{algorithm}
\usepackage{array}
\usepackage{subfig}
\usepackage[noend]{algpseudocode}
\def\BibTeX{{\rm B\kern-.05em{\sc i\kern-.025em b}\kern-.08em
    T\kern-.1667em\lower.7ex\hbox{E}\kern-.125emX}}
\begin{document}
\title{Patient-specific virtual spine straightening and vertebra inpainting: An automatic framework for osteoplasty planning}

\author{Christina Bukas, Bailiang Jian, Luis F. Rodriguez Venegas, Francesca De Benetti, Sebastian Ruehling, Anjany Sekuboyina, Jens Gempt, Jan S. Kirschke, Marie Piraud, Johannes Oberreuter, Nassir Navab and Thomas Wendler \thanks{C. Bukas and M. Piraud with Helmholtz AI, Helmholtz Zentrum Muenchen, Munich, Germany.} \thanks{C. Bukas, B. Jian, L. F. Rodriguez Venegas, F. De Benetti, N. Navab and T. Wendler are with the Chair for Computer Aided Medical Procedures and Augmented Reality, Technische Universitaet Muenchen, Garching, Germany.}\thanks{S. Ruehling, A. Sekuboyina and J. S. Kirschke are with the Department of Neuroradiology, Technische Universitaet Muenchen, Munich, Germany}\thanks{A. Sekuboyina is with the Image-Based Biomedical Modeling Group, Technische Universitaet Muenchen, Germany.}\thanks{J. Gempt is with the Department of Neurosurgery, Technische Universitaet Muenchen, Munich, Germany}\thanks{J. Oberreuter is with Reply SpA, Munich, Germany.}\thanks{N. Navab is with Computer Aided Medical Procedures, Johns Hopkins University, Baltimore, MD, USA.}}

\maketitle

\begin{abstract}
Symptomatic spinal vertebral compression fractures (VCFs) often require osteoplasty treatment. A cement-like material is injected into the bone to stabilize the fracture, restore the vertebral body height and alleviate pain. Leakage is a common complication and may occur due to too much cement being injected. In this work, we propose an automated patient-specific framework that can allow physicians to calculate an upper bound of cement for the injection and estimate the optimal outcome of osteoplasty. The framework uses the patient CT scan and the fractured vertebra label to build a virtual healthy spine using a high-level approach. Firstly, the fractured spine is segmented with a three-step Convolution Neural Network (CNN) architecture. Next, a per-vertebra rigid registration to a healthy spine atlas restores its curvature. Finally, a GAN-based inpainting approach replaces the fractured vertebra with an estimation of its original shape. Based on this outcome, we then estimate the maximum amount of bone cement for injection. We evaluate our framework by comparing the virtual vertebrae volumes of ten patients to their healthy equivalent and report an average error of 3.88$\pm$7.63\%. The presented pipeline offers a first approach to a personalized automatic high-level framework for planning osteoplasty procedures.
\end{abstract}


\section{Introduction}
This work proposes improvements in osteoplasty by image analysis using deep learning. Spinal vertebra compression fractures (VCFs) are a painful and debilitating injury to the skeleton. The main reasons for such fractures are osteoporosis, trauma or tumors, as many cancers metastasize into the vertebral column. Patients with osteoporotic fractures can be relieved from their pain and regain mobility by stabilizing the affected vertebrae.

Osteoplasty, namely kyphoplasty or vertebroplasty, is an operative procedure, during which holes are drilled or hammered from the back of the patient through the pedicles of a vertebra to the vertebral body~\cite{filippiadis2017percutaneous}.
Subsequently, the vertebral body is filled with a bone cement, which stabilizes the vertebra and relieves the patient from the pain.

Yet, it is difficult to determine the correct amount of cement to be injected~\cite{janssen2017risk}. In particular, if it is too much bone cement a leakage may put pressure on the spinal cord and can even lead to pulmonary cement embolisms~\cite{sorensen2019vertebroplasty}.

The goal of this work is to provide a virtual vertebra reconstruction in a personalized manner, taking into account patient anatomy and the vertebra type. The only data readily available comes from CT imaging, which is solely used to derive the shape of a healthy-looking vertebra, matching the patient's spine, and an upper bound for the bone cement to be injected. To extract all the information necessary we propose an automated framework based on deep learning techniques with the following steps:

\begin{enumerate}
\item \emph{Vertebrae Segmentation}: Every individual vertebra in the input CT image of a patient is detected, labeled with its anatomic denomination, and segmented. The segmentation masks serve as an input to the next step.
\item \emph{Virtual Spine Straightening}: This step simulates the restoration of the vertebral column to a healthy state after stabilizing the fractured vertebra. Usually, only the post-fracture CT image is available for a patient. To estimate the size of the restored fractured vertebra, the healthy vertebrae are compared to a spine atlas, which is scaled to match the patient. The label of the fractured vertebra is required as an input. The healthy vertebrae of the patient are registered to the atlas in a vertebra-wise rigid and deformable approach, providing a physiologically healthy vertebral column.
\item \emph{Vertebra Inpainting}: The shape of the vertebra before the fracture is estimated and its volume measured. A generative adversarial neural network (GAN) produces realistic 3D shapes of vertebrae, which are put in place of the fractured vertebra. The upper bound on the amount of cement can be estimated from the difference in volume between the inpainted and the fractured vertebra. 
\end{enumerate}
 
\begin{figure*}
    \centering
    \includegraphics[width=0.95\textwidth]{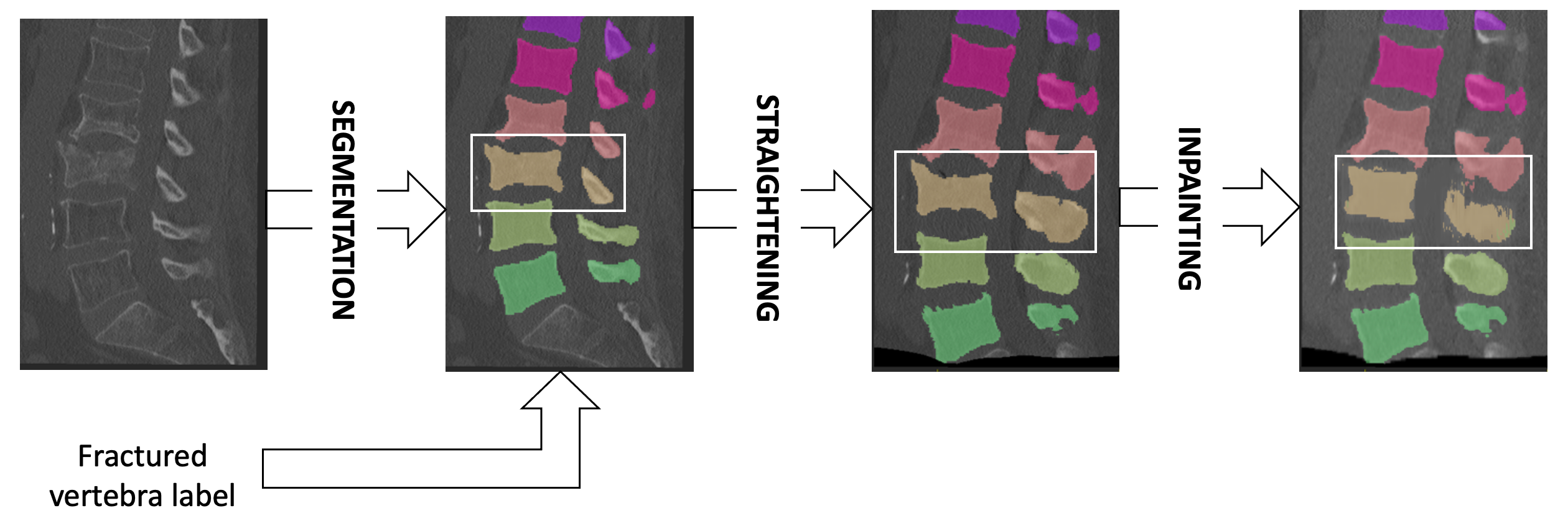}
    \caption{A visual example of the inputs and outputs of each stage of the framework.}
    \label{fig:workflow}
\end{figure*}

\begin{figure*}
\centering
\input{workflow_short_new.tex}
\caption{Proposed framework: The CT image is first used to generate a segmentation mask, which is then fed into the  spine straightening block, together with the CT image and the fractured vertebra label (e.g. "L2"). The straightened spine and corresponding segmentation are
then sent to the inpainting block to generate the final virtually healthy CT
and the maximum cement needed.}
\label{fig:workflow_visual}
\end{figure*}
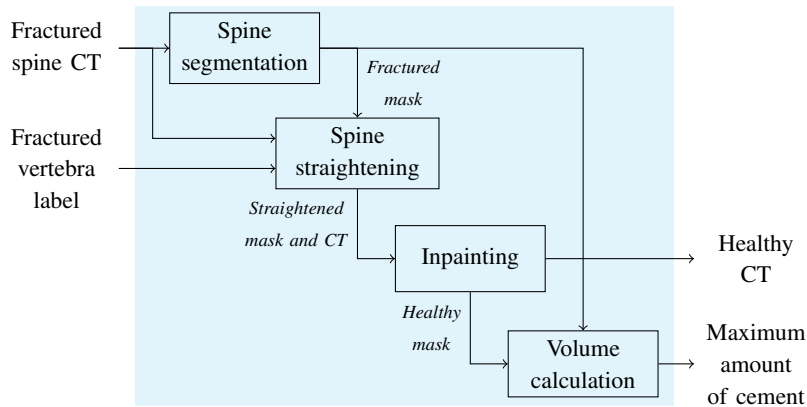

A visualization of the workflow is given in Figs.~\ref{fig:workflow} and \ref{fig:workflow_visual}.

The goal of this work is to provide a virtual vertebra reconstruction in a personalized manner, taking into account patient anatomy and the type of vertebra. The only data readily available comes from CT imaging, which is solely used to derive the shape of a healthy-looking vertebra matching the spine of the patient and an upper bound for the bone cement to be injected. To extract all the information necessary we propose an automated framework based on deep learning techniques following the workflow of Figs.~\ref{fig:workflow} and \ref{fig:workflow_visual}.

The following is a summary of the main contributions in this paper:
\begin{itemize}[leftmargin=*]
\item To our knowledge, this is the first framework to do an end-to-end spine straightening and vertebra reconstruction to estimate the upper bound of bone cement for osteoplasty. Further, the proposed pipeline is fully automatic.
\item We introduce a robust method for virtually straightening a fractured spine solely using a CT image and the label of the damaged vertebra and validate it on clinical data.
\item We propose a patient-tailored 2.5D inpainting method to generate a healthy-looking version of a fractured vertebra and its segmentation mask.
\end{itemize}

\section{Methodology}

Each step of the pipeline (Figs.~\ref{fig:workflow} and \ref{fig:workflow_visual}) is detailed in the following subsections. The models were implemented in PyTorch and their weights, along with code for running the pipeline will become publicly available upon acceptance.

\paragraph{Spine Segmentation.}
The goal of this step is to automatically generate voxel-level masks of the vertebrae. Several methods have been proposed for segmenting the spine. For instance, in the VerSe’19 challenge, eleven fully automated algorithms were benchmarked for this task~\cite{sekuboyina2020verse}. Here, we reimplement the approach of Sekuboyina et al. splitting the task into three modules: spine detection, vertebrae labeling, and vertebrae segmentation~\cite{sekuboyina2020verse,btrfly-sekuboyina}.

For detecting the spine, we employ a variant of the U-Net~\cite{ronneberger2015u,roy2018concurrent}, to regress a coarse 3D Gaussian heatmap of the spinal centerline and a 3D bounding box around it to localize the spine. For the labeling module, we use a self-implemented modified version of the BtrflyNet~\cite{btrfly-sekuboyina}, which works on 2D sagittal and coronal maximum intensity projections (MIP). We extract the MIPs from the localized spine region from the previous stage, thereby removing occlusions from ribs and pelvic bones. The network then predicts refined Gaussian heatmaps centered at each vertebra. Finally, once the vertebrae are labeled, their segmentation is modeled as a binary segmentation task, using again a modified U-Net. The spine image is first cropped to a 3D patch around each vertebral centroid previously detected. Similarly, the predicted heatmap image of the same centroid is also cropped. Both cropped vertebral image and vertebral heatmap are then fed into the network in order to segment only the vertebra of interest.

\paragraph{Virtual Spine Straightening.}

For this task, Forsberg et al. proposed  splitting registered sub-volumes of the patient's spine and atlas with non-rigid transformations, which evidently ignored the rigid nature of the vertebrae~\cite{forsberg2015atlas}. Drobny et al. proposed registering the spine with a poly-rigid transformation model thus ensuring a rigid transformation of the vertebrae, but evaluating their approach only on synthetic data~\cite{drobny2020towards}. We follow a similar approach, however tackle the issue of varying patient sizes and use a simpler method for deforming non-vertebra voxels. From the previous step, we obtain the vertebrae centroids and segmentation mask. First, we scale a spine atlas to be the same height as the patient, where the scaling factor is computed by comparing the sum of the distances between the centroids of all visible vertebrae on the image excluding the fracture(s). The label(s) of the fracture(s) is taken as an input. We calculate the distance using the first two principal components of the centroid's coordinates to mitigate the deviation resulting from the patient's position during the CT acquisition. A set of displacement fields is generated from a per-vertebra rigid registration on the segmentation masks. Next, a distance map of each vertebra is calculated, where the voxel value represents the physical distance to the nearest vertebra. The displacement fields are then compounded with Alg.~\ref{alg:displacement}, based on our biomechanic assumption that the soft tissue are transformed inversely proportional to their distance from vertebrae, similar to linear blend skinning~\cite{lewis2000}. Finally, we generate the straightened spine by resampling using the combined displacement field.

\begin{algorithm}
\caption{Algorithm used for combining the vertebrae displacement fields.}\label{alg:displacement}
\begin{algorithmic}[1]
\Require{Vertebrae set $\mathbf{V}=\{1,2,...,24\}$ excluding the fractured vertebra(e), set of displacement fields $\{\mathbf{F_i}\}_{i\in\mathbf{V}}$, $\mathbf{F_i}\in\mathbb{R}^{h\times w\times d}$, set of distance maps, $\{\mathbf{D_i}\}_{i\in\mathbf{V}}$ , $\mathbf{D_i}\in \mathbb{R}^{h\times w\times d}$, where h, w, d are height, width and depth}{}
\For{every voxel p in scan}
\If{voxel p is in vertebrae i}
\State{$\widetilde{\mathbf{F}}(p)\gets\mathbf{F_i}(p)$}
\Else{}
\State{$\widetilde{\mathbf{F}}(p)\gets \frac{\sum_j \mathbf{D_j}^{-1}(p)*\mathbf{F_j}(p)}{\sum_j\mathbf{D_j}^{-1}(p)}$}
\EndIf
\EndFor
\Return{combined displacement field $\widetilde{\mathbf{F}}$}
\end{algorithmic}
\end{algorithm}

\paragraph{Vertebra Inpainting.}

After spine straightening, the next task in the framework is to replace the fractured vertebra with its healthy equivalent. This is done using inpainting, which has previously been used in medicine for predicting missing information~\cite{armanious2019adversarial,torrado2021inpainting}, removing lesions~\cite{zhang2020robust} or correcting limited-angle acquisitions~\cite{zhao2018unsupervised,li2019promising,wang2020effective}. None of these works tackles a similar problem to ours. Here, though we are dealing with 3D volumes, when focusing on vertebral bodies, the sagittal and coronal views are expected to provide sufficient information for reconstruction. We, therefore, chose to apply two 2D models to a volume, one trained on sagittal slices, the other on coronal, and fuse the outputs. While we are mainly interested in the segmentation mask for the framework pipeline, we tackle the inpainting as a multi-task learning problem, and train a model to generate the inpainted CT image as well as the segmentation mask. This enforces the model to better learn the training data distribution. 

We remove the fractured vertebra from the CT and segmentation mask slices by applying a binary mask, which is generated from the segmentation and fractured label. These are the inputs to the inpainting network, whose architecture extends that of Yu et al.~\cite{yu-context-att}. We adjust their architecture here to receive two inputs, the corrupted CT slice, and corresponding segmentation. Similarly to the original authors' network, the input first passes through a coarse generator for a first estimate of the inpainted result and next through a refinement network, both of which have a U-like architecture. The last layer of the refinement network has been adjusted to two final parallel layers, which output the inpainted image and the inpainted mask. Four discriminators follow the generator; two, for image and mask, local for the patch region and two global, for the entire image space. We use the same losses as presented in \cite{yu-context-att}, extending the generator loss by adding dice losses for the patch region and the background of the mask and adjusting the total discriminator loss to accommodate all four sub-networks. To obtain the final volumes, we apply a simple averaging of the intensities in the two CT volumes, while for the segmentation, the softmax outputs of the segmentation layer are averaged before acquiring the final predictions. From this we then compute the virtually healthy vertebra volume, and by substracting from this the original fracture vertebra volume, compute an upper bound for the cement.

\section{Experimental Setup}

For training and evaluating the individual steps of our framework, as well as evaluating the overall pipeline we use three separate datasets.

\paragraph{Segmentation Training Dataset.}
For the vertebra segmentation pipeline, we used the public dataset VerSe’20~\cite{loffler2020vertebral,sekuboyina2020labeling,sekuboyina2020verse}, which consists of 100 patients with ground-truth vertebrae centroid annotations and segmentation masks. This data comes from multiple multi-detector CT scanners, and includes a wide range of fields-of-view, scan settings, and certain cases with vertebral fractures, metallic implants, and foreign materials. We randomly split this dataset into a training set of 70 patients, and a validation and testing set of 15 patients, respectively.

\paragraph{Inpainting Training Dataset.}
Since we want our model to learn to inpaint the region of interest with a healthy vertebra, we use a healthy dataset for training. We created two 2D datasets, one of coronal and one of sagittal views, from a total of 110 volumes (95 confirmed healthy spines from our institution and 15 from the CSI challenge\footnote{http://csi-workshop.weebly.com/challenges.html} which we verified to be healthy). From these volumes ten were set aside for testing and ten for validation of both networks. Depending on the size and resolutions of the scans, the number of slices and spine regions taken from each volume varied, making up a total of 3557 sagittal and 1358 coronal images, each consisting of five vertebra.

\paragraph{Dataset for Validation of Virtual Spine Straightening and Overall Pipeline.}
For the evaluation of the straightening and the overall workflow, we obtained a list of 316 patients who underwent kyphoplasty between 2014 and 2019 at our institution. We filtered these to include only patients who (1) have the pre-fracture, pre-operative and post-operative CT scans, and (2) have CT images with voxel sizes smaller than 1 mm in the x and y direction and smaller than 3 mm on z axis. After applying these criteria, our dataset included ten patients. For the sake of completeness, some of these ten patients have more than one fractured vertebra, which enabled us to evaluate the pipeline on a total of 15 vertebrae. The CT scans included in the dataset have resolution between 0.28 and 0.97 mm in the x and y direction, and between 0.7 and 3.0 mm in the z direction.

\paragraph{Segmentation Pipeline Training.}
We trained each module of the segmentation pipeline independently, following the same dataset split for every task. For the detection stage, we first built the ground-truth spine heatmap by combining the individual heatmaps of the vertebrae landmarks. We then trained the 3D U-Net variant with the L2 loss, using the Adam optimizer, a batch size of two and a learning rate of 1e-3 until convergence (77 epochs). Next, we extracted the bounding box around the predicted spine heatmap. For the labelling stage we trained the modified BtrflyNet on the MIPs extracted from the previous stage bounding box. Here, we used the same hyperparameters and L2 loss, since this task is also modeled as a heatmap regression, and trained until convergence (280 epochs). Finally, we used the vertebral patches and heatmaps from the previous stage to train the modified  3D U-Net.  We used the Dice Loss and trained for 25 epochs, again with the same hyperparameters. 

\paragraph{Inpainting Model Training.}
We trained the two models on the lateral and coronal datasets, where each slice contains four visible vertebrae and one digitally erased. Thus, each image was sent five times through the networks, which were trained using a batch size of 16, with Adam for the optimization and initial learning rates of 0.001 and 1e-4 for the generator and discriminator. A grid search was applied to find optimal loss weights and learning rates, in the space around the original authors' choice. We implement early stopping considering the segmentation metrics, since for our pipeline the segmentation result is of higher relevance. The best performing models on the validation sets were chosen.

\section{Results}

\paragraph{Evaluation of Segmentation.}
For assessing the performance of the vertebrae segmentation framework, we employed the dice coefficient metric. We computed this metric at a vertebra level over all the vertebrae annotated in the ground truth from the VerSe2020 mask labels. The average dice score over all patients in the test set is 87.06$\pm$9.54\%, a value well in the range of the reported values from the literature (83.06 to 93.01\%, Tab.~\ref{tab:inpaiting}, e.g., \cite{sekuboyina2020verse}).

\begin{table}
\centering
\caption{Evaluation of Vertebra Segmentation as compared to state-of-the-art methods}
\label{tab:inpaiting}
\begin{tabular}{|l|l|}
\hline
Team &  Dice \\
\hline
Payer C. & 90.90  \\
Lessmann N. & 85.08  \\
Sekuboyina A. & 83.06  \\
Chen M. & 93.011  \\
Hu Y. & 84.07  \\
\textbf{Ours} & 87.06  \\
\hline
\end{tabular}
\end{table}

\begin{figure}
\centering
\includegraphics[width=.45\textwidth]{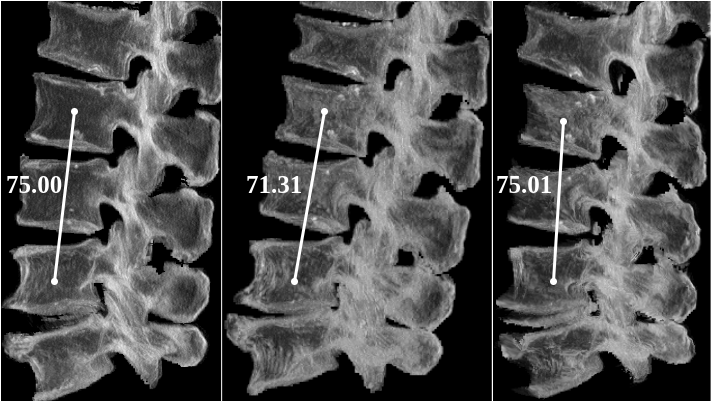}
\caption{Visual results of the straightening step. From left to right: MIPs of the pre-fractured spine, post-fractured and straightened spine, with the distance (in mm) between the vertebrae above and below the fractured one.}
\label{fig:straightening_example}
\end{figure}

\begin{figure}
\centering
\subfloat[\emph{Inpainting}]
   {\includegraphics[width=.4\textwidth]{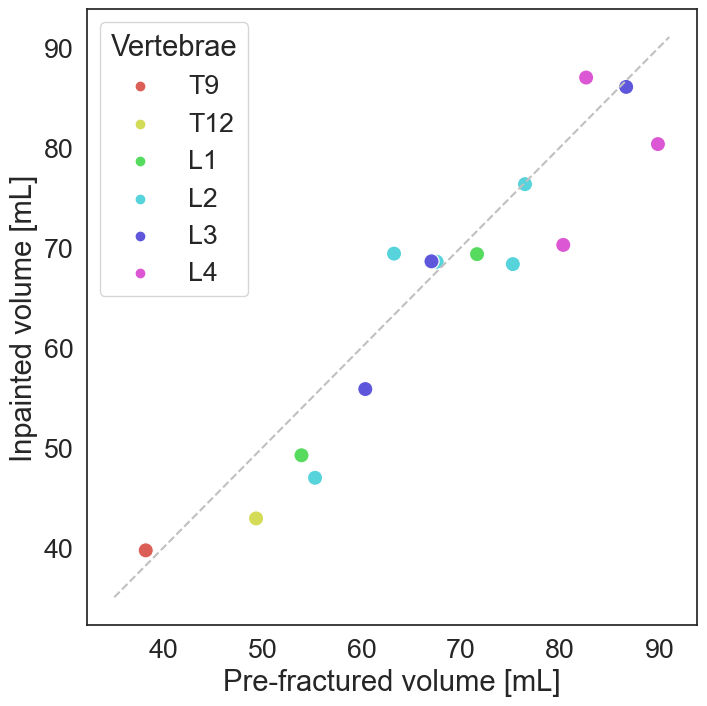}} \quad \quad \quad
\subfloat[\emph{Straightening}]
   {\includegraphics[width=.4\textwidth]{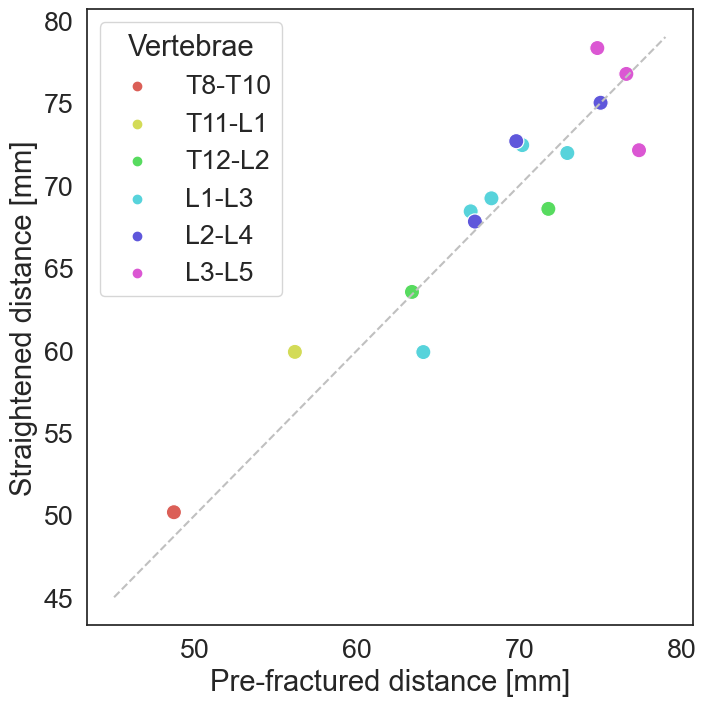}}
\caption{Quantitative results of the inpainting and straightening steps.} 

\label{fig:straightening_and_overall}
\end{figure}

\paragraph{Evaluation of Virtual Spine Straightening.}
We evaluated the spine straightening algorithm by utilizing what we call the \emph{fracture distance}. This is the physical distance between the vertebrae above and below the fracture, which should increase to provide enough space for the inpainting of a healthy vertebra. In order to make the distance comparable, the pre-fracture spine CT was also registered to the atlas space using the straightening algorithm. We therefore compared the fracture distance of straightened pre-fracture, raw post-fracture and straightened post-fracture spine of the patients (Figs.~\ref{fig:straightening_example} and \ref{fig:straightening_and_overall}). The average error on the \emph{fracture distance} yielded 0.23$\pm$2.68 mm, or equivalently, 0.50$\pm$3.95\%. The per-vertebra results are tabulated in Tab.~\ref{tab:straightening}.

\begin{table*}[]
\caption{Results of the straightening of the spine. The values in the table is the distance between the two vertebrae specified in the second column in the pre-fractured, the fractured and straightened spine CT. The pre-fractured and and straightened distances are computed on the corresponding scaled CT scans, but the fractured distance is computed on the original (therefore not scaled) CT scan. The error (between the straightened and pre-fractured distances) and the relative error are also reported.\\}
\label{tab:straightening}
    \centering
\begin{tabular}{|c|c|c|c|c|c|c|}
\hline
Patient & Vertebrae&Pre-fractured &Fractured &Straightened &Error&RE\\
&&distance [mm]&distance [mm]& distance [mm]&[mm]&[\%]\\
\hline
1&T12-L2&71.78&57.28&68.57&-3.21&-4.54\\
&L2-L4&75.00&71.31&75.01&0.02&0.02\\
&L3-L5&76.59&70.45&76.76&0.17&0.23\\
2&T8-T10&48.69&48.48&50.16&1.47&3.02\\
3&T11-L1&56.15&50.25&59.89&3.74&6.66\\
4&L1-L3&64.07&59.23&59.88&-4.19&-6.39\\
&L2-L4&67.25&62.53&67.80&0.55&0.82\\
5&L1-L3&66.99&69.07&68.42&1.44&2.02\\
6&L1-L3&68.27&60.31&69.21&0.94&1.37\\
&L2-L4&69.80&64.19&72.68&2.88&4.21\\
7&L1-L3&72.95&71.68&71.96&-1.00&-1.33\\
&L3-L5&77.37&73.27&72.13&-5.24&-6.86\\
8&L3-L5&74.80&76.27&78.33&3.53&4.77\\
9&L1-L3&70.17&68.54&72.44&2.27&3.30\\
10&T12-L2&63.37&64.64&63.53&0.16&0.25\\
\hline
\textbf{Average}&&&&&\textbf{0.23}&\textbf{0.50}\\
\textbf{STD}&&&&&\textbf{2.68}&\textbf{3.95}\\
\hline
\end{tabular}
\end{table*}

\begin{figure*}
\centering
\subfloat
   {\includegraphics[width=.45\textwidth]{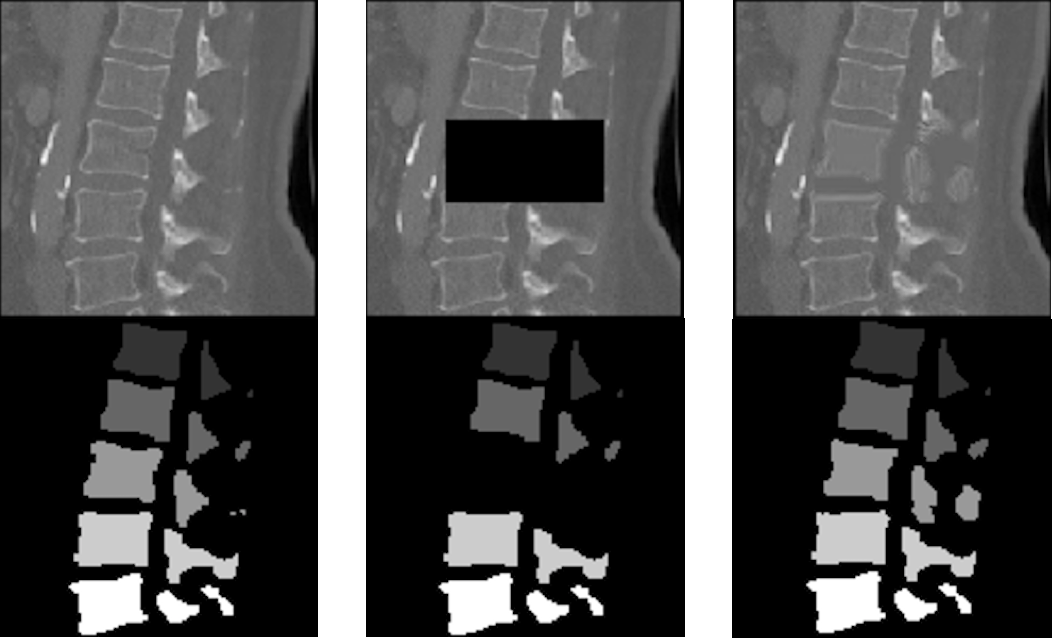}} \quad \quad \quad
\subfloat
   {\includegraphics[width=.45\textwidth]{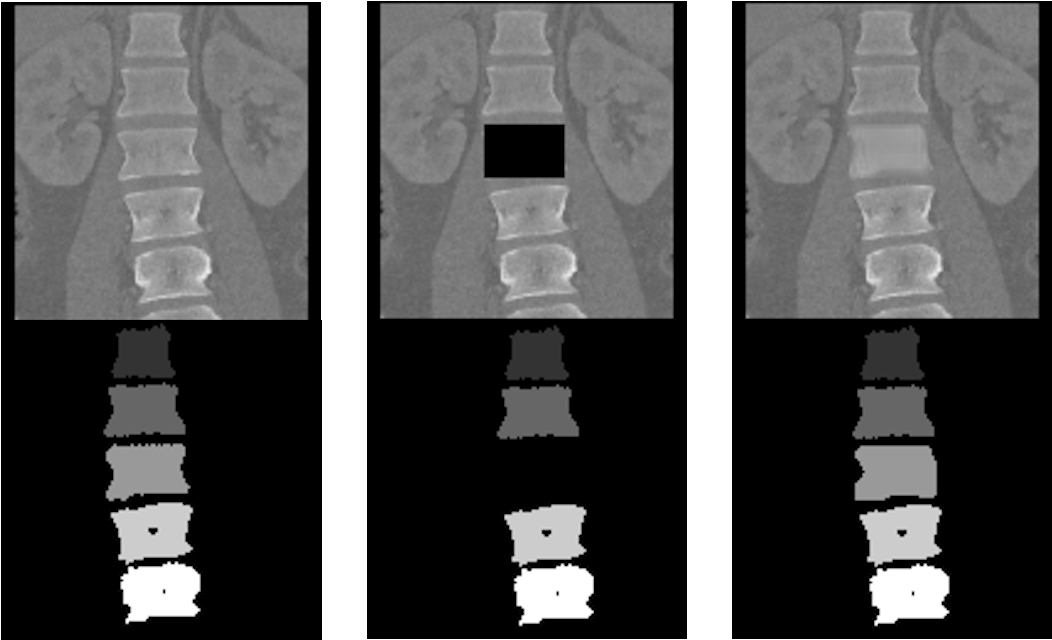}}
\caption{Visual results of the inpainting step: From left to right: ground-truth, input and output of the sagittal (left pane) and coronal (right pane) GANs. The first row shows the CTs, while the second depicts the segmentation masks.}
\label{fig:inpaiting_example}
\end{figure*}

\paragraph{Evaluation of Inpainting.}
To evaluate the results of the inpainting model we computed the structural similarity index metric (SSIM) and peak signal-to-noise Ratio (PSNR) of the output CT, as well as the intersection over union (IoU) of the segmentation in the patch region of interest and the mean relative volume error (MRE) of the said vertebra. Fusing the coronal and sagittal outputs improves the individual results, giving our final model an SSIM of 0.82 and PSNR of 26.45 dB for the image, an IoU of 0.76 for the mask and an MRE of 19\% for the vertebra volume (see Fig.~\ref{fig:inpaiting_example} and Tab.~\ref{tab:inpaiting}).

\begin{table}
\centering
\caption{Evaluation of inpainting per view. The metrics used for the CT inpainting are structural similarity index metric (SSIM) and peak signal-to-noise Ratio (PSNR), while for the segmentation masks we \\}
\label{tab:inpaiting}
\begin{tabular}{|l|c|c|c|c|}
\hline
Method &  Image SSIM & Image PSNR & Mask MRE & Mask IoU\\
\hline
Sagittal & 0.771 & 25.380 & 0.228 & 0.734 \\
Coronal & \textbf{0.830} & 26.448  & 0.316 & 0.691 \\
Fusion & 0.823 &\textbf{28.16}  & \textbf{0.190} & \textbf{0.757} \\
\hline
\end{tabular}
\end{table}

\paragraph{Evaluation of Overall Pipeline.}
We evaluated the performance of our framework in an end-to-end manner by using ten patients. We note again here, that for these patients a CT scan before and after the VCF were fortuitously available. To evaluate the effectiveness of our pipeline, we compare the volume of the vertebrae before the VCF (pre-fractured) with the output of our framework, i.e., the virtually straightened inpainted post-fracture scan. Overall, our method has an error of 2.61$\pm$5.07 mL, which translates to a relative error of 3.08$\pm$7.63 \%. Fig.~\ref{fig:straightening_and_overall} shows the correlation between the pre-fracture and straightened inpainted vertebra volumes. The table with the results at a per-vertebra level is included in the supplementary material.

\begin{table*}[]
\caption{Framework analysis}
\label{tab:volume_evaluation}
\centering
\begin{tabular}{|c|c|c|c|c|p{10mm}<{\centering}|p{10mm}<{\centering}|}
\hline
Patient & Vertebra & Pre-fractured & Fractured & Inpainted & Error & RE \\
&&volume [mL]&volume [mL]&volume [mL]&[mL]&[\%]\\
\hline
1&L1&71.61&62.41&69.30& 2.31& 3.23\\
&L3&86.65&79.89&86.01& 0.64& 0.74\\
&L4&82.61&79.65&86.95&-4.34&-5.25\\
2&T9&38.18&37.69&39.69&-1.51&-3.96\\
3&T12&49.31&38.99&42.88&6.43&13.05\\
4&L2&63.23&58.90&69.35&-6.12&-9.67\\
&L3&67.01&66.21&68.57&-1.57&-2.34\\
5&L2&76.45&72.05&76.29&0.16&0.21\\
6&L2&55.26&44.02&46.94&8.32&15.05\\
6&L3&60.33&55.35&55.81&4.52&7.50\\
7&L2&75.22&69.32&68.30&6.92&9.20\\
&L4&80.30&73.08&70.22&10.09&12.56\\
8&L4&89.86&86.41&80.30&9.56&10.64\\
9&L2&67.53&64.84&68.54&-1.01&-1.49\\
10&L1&53.89&51.66&49.19&4.70&8.72\\
\hline
\textbf{Average}&&&&&\textbf{2.61}&\textbf{3.88}\\
\textbf{STD} &&&&&\textbf{5.07}&\textbf{7.63}\\
\hline
\end{tabular}
\end{table*}

\paragraph{Ablation study.}
A final relevant evaluation is that of validating the significance of the spine straightening to the inpainting results. For this we compute the volume of the inpainted vertebra directly using the fractured spine, without using the straightening algorithm for the same ten patients (15 vertebrae). We compare here the volume of the inpainted vertebrae with and without straightening (Tab.~\ref{tab:ablation}).

\begin{table*}[]
\centering
\caption{Results of the ablation study in which the straightening of the spine was omitted. The values summarize the effect of the spine straightening in the pipeline, by reporting the error (between the inpainted and pre-fractured distances) and the relative error.}
\label{tab:ablation}
\centering
\begin{tabular}{|c|c|c|c|p{10mm}<{\centering}|p{10mm}<{\centering}|c|c|p{10mm}<{\centering}|p{10mm}<{\centering}|}
\hline
& & Pre-fractured & \multicolumn{3}{c|}{Straightening} & \multicolumn{3}{c|}{W/o straightening} \\
\cline{4-9}
Patient & Vertebra&volume&Inpainted & Error & RE & Inpainted & Error & RE\\
&&[mL]&volume [mL]&[mL]&[\%]&volume [mL]&[mL]&[\%]\\
\hline
1&L1&71.61&69.30&2.31&3.23&55.462&16.14&22.55\\
&L3&86.65&86.01&0.64&0.74&74.14&12.51&14.44\\
&L4&82.61&86.95&-4.34&-5.25&71.11&11.51&13.93\\
2&T9&38.18&39.69&-1.51&-3.96&33.76&4.42&11.58\\
3&T12&49.31&42.88&6.43&13.05&31.44&17.87&36.25\\
4&L2&63.23&69.35&-6.12&-9.67&55.02&8.21&12.99\\
&L3&67.01&68.57&-1.57&-2.34&59.02&7.99&11.92\\
5&L2&76.45&76.29&0.16&0.21&61.74&14.70&19.23\\
6&L2&55.26&46.94&8.32&15.05&42.64&12.62&22.83\\
6&L3&60.33&55.81&4.52&7.50&51.96&8.37&13.87\\
7&L2&75.22&68.30&6.92&9.20&59.45&15.77&20.96\\
&L4&80.30&70.22&10.09&12.56&66.10&14.20&17.69\\
8&L4&89.86&80.30&9.56&10.64&79.43&10.43&11.61\\
9&L2&67.53&68.54&-1.01&-1.49&65.04&2.49&3.69\\
10&L1&53.89&49.19&4.70&8.72&47.47&6.42&11.91\\
\hline
\textbf{Average}&&&&\textbf{2.61}&\textbf{3.88}&&\textbf{10.91}&\textbf{16.08}\\
\textbf{STD}&&&&\textbf{5.07}&\textbf{7.63}&&\textbf{4.52}&\textbf{7.46}\\
\hline
\end{tabular}
\end{table*}

\section{Discussion}

In this work, we have presented an integrated framework to estimate a realistic healthy state for fractured vertebrae. Using only a patient's post-traumatic CT image, a healthy vertebra shape replaces the fractured one, virtually restoring the spine. Every single stage in this pipeline is necessarily required to obtain a genuine result. As a clinical application of our work, we envision that the volume of the estimated vertebra can be used to estimate the amount of bone cement needed to stabilize the vertebra without leakage, albeit we are not performing this very analysis. Therefore, we view the estimated volume as an upper bound on the injection of material. 

As the primary output of the workflow is a 2.5D reconstructed image generated by a GAN trained on healthy spines, it is necessary that the healthy vertebrae in the image to be inpainted are resembling a healthy spine and provide accurate space for the inpainting. The virtual spine straightening step solves this problem.

We have also validated the distance between the vertebrae neighbouring the fracture for those patients, where pre-traumatic imaging is available.

We validate the results of the inpainting on the same set of patients showing that it tallies very well with their pre-fractured state at a relative error of below 4\%. In general, the method is slightly underestimating the volume of the vertebrae, which can be considered conservative given the necessity for an upper bound. Admittedly, this was not engineered and will be investigated further.

In its current state, our framework demonstrates its high potential as an automatic method for reconstruction of healthy-looking medical images, requiring minimal diagnostic input. The reported ability to derive quantitative results proves the usefulness of deep learning approaches for planning interventions.

\section*{Acknowledgements}
The authors would like to thank the support of Magdalini Paschali, Stefan Walke, Sebastian Lutz and the rest of the team at the Interdisciplinary Research Lab at Klinikum rechts der Isar.

\bibliographystyle{splncs04}
\bibliography{paper.bib}

\end{document}

%% file: workflow_short_new.tex

\begin{tikzpicture}[scale=1.0, transform shape]

\def \blocksize {25};

\draw[cyan!10, fill=cyan!10] (0.55,0.55) -- (7.7,0.55) -- (7.7,-4.75) -- (0.55,-4.75) -- cycle;

\node[text width=40, align=center](in1) at (-0.5,0) {\small{Fractured spine CT}};
\node[text width=40, align=center](in2) at (-0.5,-1.6) {\small{Fractured vertebra label}};
\node[draw,minimum height=\blocksize,text width=50, align=center](step_1) at (2,0) {\small{Spine segmentation}};
\node[draw,minimum height=\blocksize,text width=55, align=center](step_2) at (3.5,-1.4) {\small{Spine straightening}};
\node[draw,minimum height=\blocksize,text width=50, align=center](step_3) at (5,-2.8) {\small{Inpainting}};
\node[minimum height=\blocksize,text width=40, align=center](out_1) at (8.8,-2.8) {\small{Healthy CT}};
\node[draw,minimum height=\blocksize,text width=50, align=center](step_5) at (6.5,-4.2) {\small{Volume calculation}};
\node[minimum height=\blocksize,text width=40, align=center](out) at (8.8,-4.2) {\small{Maximum amount of cement}};

\draw[->](in1) -- (step_1);
\draw[->](in1.east) -|++ (0.45cm,-1.2cm) -- ($(step_2.west)+(0,0.2)$);
\draw[->](in2) -- ($(step_2.west)-(0,0.2)$);
\draw[->]($(step_1.east)$) -|  ($(step_2.north)$) node[near end, right, align=center]{\textit{\scriptsize{Fractured}}\\ \textit{\scriptsize{mask}}};
\draw[->](step_2) |- (step_3) node[near start, left, align=center]{\textit{\scriptsize{Straightened}}\\ \textit{\scriptsize{mask and CT}}};
\draw[->](step_1) -| (step_5);
\draw[->](step_3.south) |- (step_5) node[near start, left, align=center]{\textit{\scriptsize{Healthy}}\\ \textit{\scriptsize{mask }}};
\draw[->](step_5) -- (out);
\draw[->](step_3.east) |- (out_1);
\end{tikzpicture}